\documentclass[10pt,twocolumn,letterpaper]{article}

\usepackage{cvpr}
\usepackage{times}
\usepackage{graphicx}
\usepackage{amsmath}
\usepackage{amssymb}
\usepackage{wasysym}
\usepackage{changepage}
\usepackage{ragged2e}
\usepackage{braket}
\usepackage{booktabs}
\usepackage{etoolbox}
\usepackage{multirow}
\usepackage{topcapt}
\usepackage{float}
\usepackage{appendix}
\usepackage{subfig}
\usepackage{lipsum}

\usepackage[breaklinks=true,bookmarks=false]{hyperref}

\cvprfinalcopy 

\def\cvprPaperID{****} 

\begin{document}

\title{GPS-Net: Graph Property Sensing Network for Scene Graph Generation}
\author{Xin Lin$^1$ \quad Changxing Ding$^{1}$  \quad Jinquan Zeng$^1$ \quad Dacheng Tao$^2$\\
$^1$ School of Electronic and Information Engineering, South China University of Technology \\
$^2$ UBTECH Sydney AI Centre, School of Computer Science, Faculty of Engineering,\\ The University of Sydney, Darlington, NSW 2008, Australia\\
{{\tt\small$\left\{ {} \right.$eelinxin,eetakchatsau$\left. {} \right\}$@mail.scut.edu.cn\quad chxding@scut.edu.cn\quad dacheng.tao@sydney.edu.au}}}
\twocolumn[{%
\maketitle
\begin{center}
\setlength\abovecaptionskip{0.2\baselineskip}
\setlength\belowcaptionskip{1pt}
    \centering
    \includegraphics[width=1\textwidth,height=7cm]{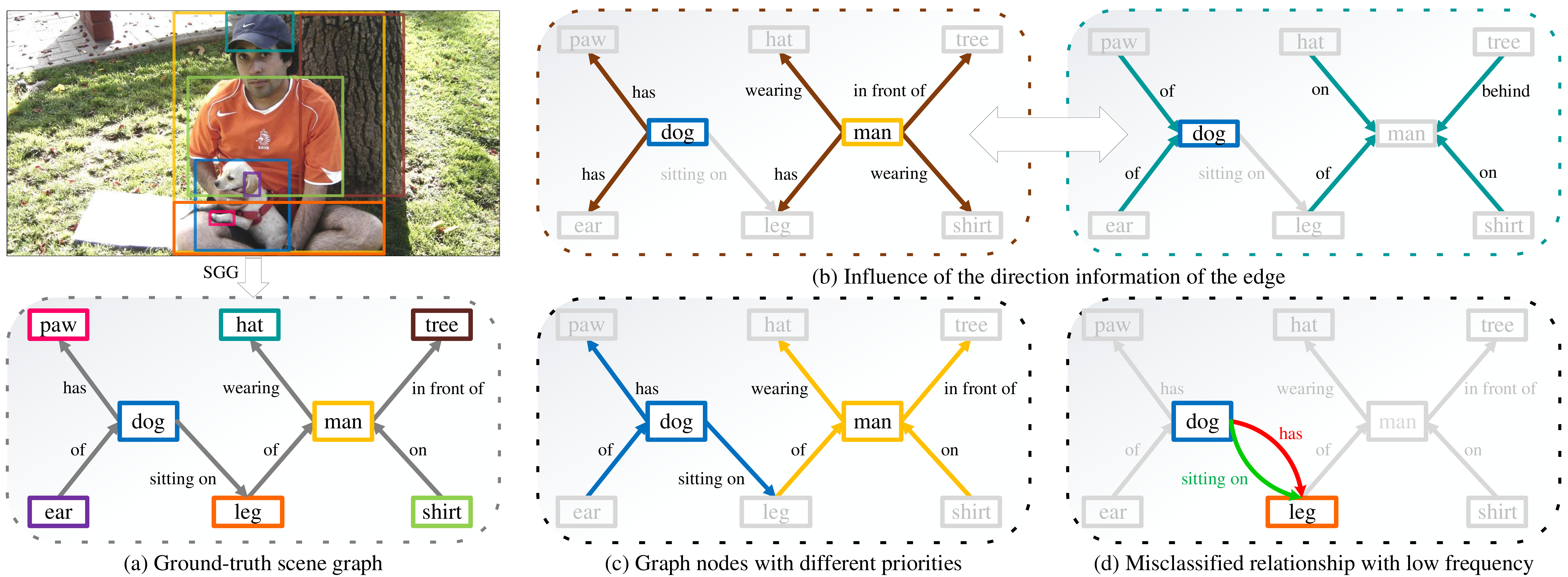}
    \captionof{figure}{(a) The ground-truth scene graph for one image. (b) The direction of the edge specifies the subject and object, and also affects the relationship type and node-specific context. (c) The priority of nodes varies, according to the number of triplets included in the graph. (d) The long-tailed distribution of relationships causes error for low-frequency relationships, \textit {e.g.}, the failure in recognizing \textit{sitting on}.}\label{example1}
\end{center}%
}]

\maketitle
\thispagestyle{empty}

\begin{abstract}
Scene graph generation (SGG) aims to detect objects in an image along with their pairwise relationships. There are three key properties of scene graph that have been underexplored in recent works: namely, the edge direction information, the difference in priority between nodes, and the long-tailed distribution of relationships. Accordingly, in this paper, we propose a Graph Property Sensing Network (GPS-Net) that fully explores these three properties for SGG. First, we propose a novel message passing module that augments the node feature with node-specific contextual information and encodes the edge direction information via a tri-linear model. Second, we introduce a node priority sensitive loss to reflect the difference in priority between nodes during training. This is achieved by designing a mapping function that adjusts the focusing parameter in the focal loss. Third, since the frequency of relationships is affected by the long-tailed distribution problem, we mitigate this issue by first softening the distribution and then enabling it to be adjusted for each subject-object pair according to their visual appearance. Systematic experiments demonstrate the effectiveness of the proposed techniques. Moreover, GPS-Net achieves state-of-the-art performance on three popular databases: VG, OI, and VRD by significant gains under various settings and metrics. The code and models are available at \url{https://github.com/taksau/GPS-Net}.
\end{abstract}

\section{Introduction}
Scene Graph Generation (SGG) provides an efficient way for scene understanding and valuable assistance for various computer vision tasks, including image captioning \cite{MSDN}, visual question answering \cite{vctree} and 3D scene synthesis \cite{3D}. This is mainly because the scene graph \cite{scenegraph} not only records the categories and locations of objects in the scene but also represents pairwise visual relationships of objects.

As illustrated in Figure \ref{example1}(a), a scene graph is composed of multiple triplets in the form $<$subject-relationship-object$>$. Specifically, an object is denoted as a node with its category label, and a relationship is characterized by a directed edge between two nodes with a specific category of predicate. The direction of the edge specifies the subject and object in a triplet. Due to the complexity in relationship characterization and the imbalanced nature of the training data, SGG has emerged as a challenging task in computer vision.

Multiple key properties of the scene graph have been under-explored in the existing research, such as  \cite{imp, contrastive, neural-motif}. The first of these is edge direction. Indeed, edge direction not only indicates the subject and object in a triplet, but also affects the class of the relationship. Besides, it influences the context information for the corresponding node, as shown in recent works \cite{gat, direction}. An example is described in Figure \ref{example1}(b), if the direction flow between \textit{man} and the other objects is reversed, the focus of the context will change and thus affects the context information for all the related nodes. This is because that the importance of nodes varies according to the number of triplets they are included in the graph. As illustrated in Figure \ref{example1}(c), \textit{leg}, \textit{dog} and \textit{man} are involved in two, three, and four triplets in the graph, respectively. Hence, considering the contribution of each node to this scene graph, the priority in object detection should follow the order: $man>dog>leg$. However, existing works usually treat all nodes equally in a scene graph.

Here, we propose a novel direction-aware message passing (DMP) module that makes use of the edge direction information. DMP enhances the feature of each node by providing node-specific contextual information with the following strategies. First, instead of using the popular first-order linear model \cite{attentive, agent}, DMP adopts a tri-linear model based on Tucker decomposition \cite{mutan} to produce an attention map that guides message passing. In the tri-linear model, the edge direction affects the attention scores  produced. Second, we augment the attention map with its transpose to account for the uncertainty of the edge direction in the message passing step. Third, a transformer layer is employed to refine the obtained contextual information.

Afterward, we devise a node priority-sensitive loss (NPS-loss) to encode the difference in priority between nodes in a scene graph. Specifically, we maneuver the loss contribution of each node by adjusting the focusing parameter of the focal loss \cite{focalloss}. This adjustment is based on the frequency of each node included in the triplets of the graph. Consequently, the network can pay more attention to high priority nodes during training. Comparing with \cite{agent} (exploiting a non-differentiable local-sensitive loss function to represent the node priority), the proposed NPS-loss is differentiable and convex, and so it can be easily optimized by gradient descent based methods and deployed to other SGG models.

Finally, the frequency distribution of relationships has proven to be useful as prior knowledge in relationship prediction \cite{neural-motif}. However, since this distribution is long-tailed, its effectiveness as the prior is largely degraded. For example, as shown in Figure \ref{example1}(d), one SGG model tends to misclassify \textit {sitting on} as \textit {has} since the occurrence rate of the latter is relatively high. Accordingly, we propose two strategies to handle this problem. First, we utilize a log-softmax function to soften the frequency distribution of relationships. Second, we propose an attention model to adaptively modify the frequency distribution for each subject-object pair according to their visual appearance.

In summary, the innovation of the proposed GPS-Net is three-fold: (1) DMP for message passing, which enhances the node feature with node-specific contextual information; (2) NPS-loss to encode the difference in priority between different nodes; and (3) a novel method for handling the long-tailed distribution of relationships. The efficacy of the proposed GPS-Net is systematically evaluated on three popular SGG databases: Visual Genome (VG) \cite{VGdataset}, OpenImages (OI) \cite{OI} and Visual Relationship Detection (VRD) \cite{vrd}. Experimental results demonstrate that the proposed GPS-Net consistently achieves top-level performance.

\begin{figure*}[tbp]
\setlength\abovecaptionskip{-1.8\baselineskip}
\setlength\belowcaptionskip{2pt}
\begin{center}
\includegraphics[scale=0.158]{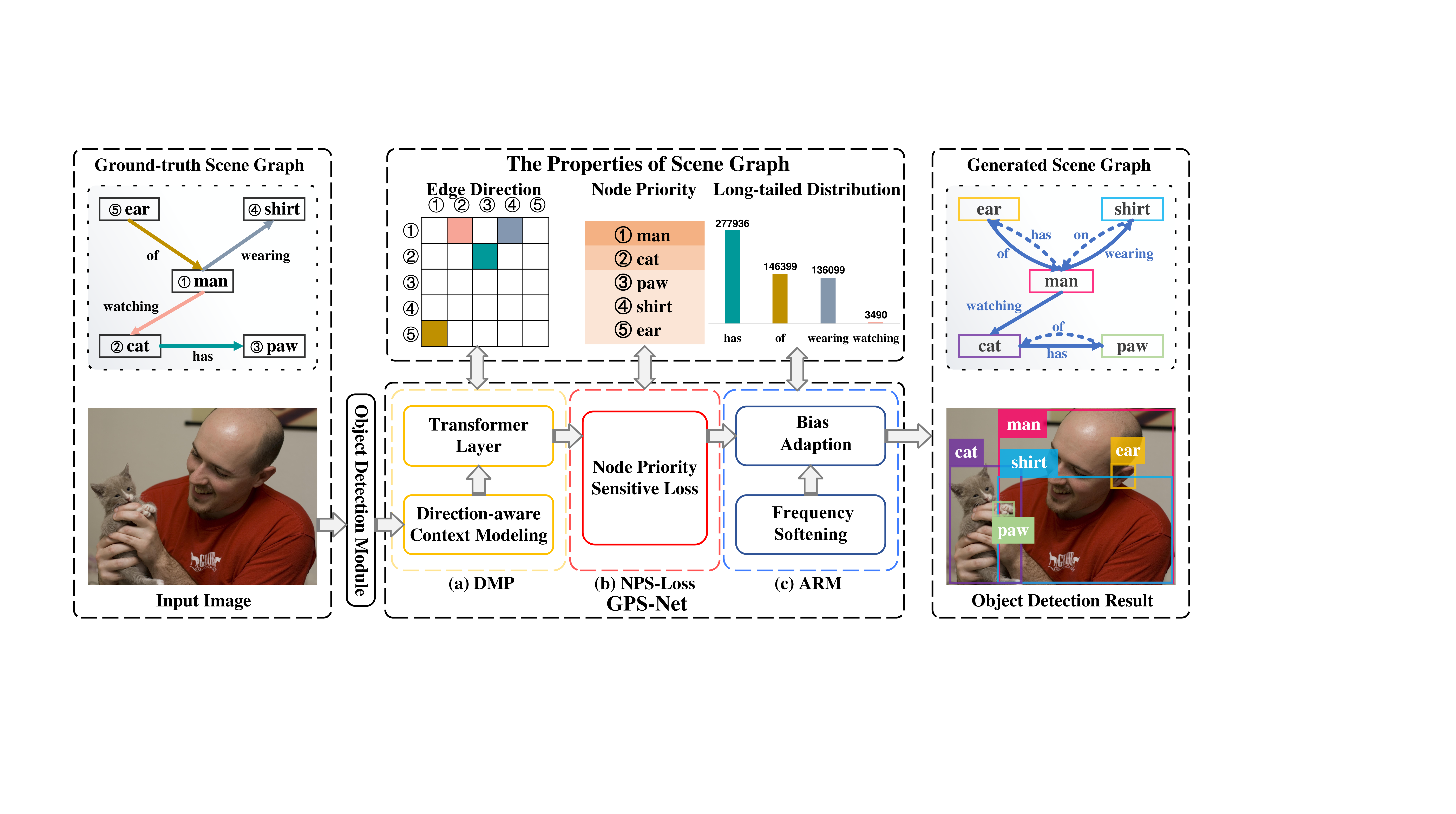}
\topcaption{The framework of GPS-Net. GPS-Net adopts Faster R-CNN to obtain the location and visual feature of object proposals. It includes three new modules for SGG: (1) a novel message passing module named DMP that enhances the node feature with node-specific contextual information; (2) a new loss function named NPS-loss that reflects the difference in priority between different nodes; (3) an adaptive reasoning module (ARM) to handle the long-tailed distribution of relationships.}
  \label{example2}
\end{center}
\end{figure*}

\section{Related Work}
\pagestyle{empty}
\textbf{Visual Context Modeling:}
Recent approaches for visual context modeling can be divided into two categories, which model the global and object-specific context, respectively. To model the global context, SENet \cite{senet} and PSANet \cite{psanet} adopt rescaling to different channels in feature maps for feature fusion. In addition, Neural Motif \cite {neural-motif} represents the global context via Long Short-term Memory Networks.

To model the object-specific context, NLNet \cite{nlnet} adopts self-attention mechanism to model the pixel-level pairwise relationships. CCNet \cite{ccnet} accelerates NLNet via stacking two criss-cross blocks. However, as pointed out in \cite {gcnet}, these methods \cite{graphrcnn, kern, gpi} may fail to learn object-specific context due to the utilization of the first-order linear model. To address this issue, we design a direction-aware message passing module to generate node-specific context via a tri-linear model.

\textbf{Scene Graph Generation}. Existing SGG approaches can be roughly divided into two categories: namely, one-stage methods and two-stage methods. Generally speaking, most one-stage methods focus on object detection and relationship representation \cite{MSDN, imp, attentive, vrd, graphrcnn, factorize}, but almost ignore the intrinsic properties of scene graphs, \textit {e.g.}, the edge direction and node priority. To further capture the attributes of scene graph, two-stage methods utilize an extra training stage to refine the results produced by the first stage training. For example, \cite{gpi} utilizes the permutation-invariant representations of scene graphs to refine the results of \cite{neural-motif}. Besides, \cite{vctree} utilizes dynamic tree structure to characterize the acyclic property of scene graph. Meanwhile, \cite{agent} adopts a graph-level metric to learn the node priority of scene graph. However, the adopted loss functions in \cite{vctree, agent} are non-differentiable and therefore hard to optimize. The proposed approach is a one-stage method but has the following advantages comparing with existing works. First, it explores the properties of the scene graph more appropriately. Second, it is easy to optimize and deploy to existing models.

\section{Approach}
Figure \ref{example2} illustrates the proposed GPS-Net. We employ Faster R-CNN \cite{fasterrcnn} to obtain object proposals for each image. We adopt exactly the same way as \cite{neural-motif} to obtain the feature for each proposal. There are $O$ object categories (including background) and $R$ relationship categories (including non-relationship). The visual feature for the $i$-th proposal is formed by concatenating the appearance features ${{\emph {\textbf{v}}}_i} \in \mathbb R^{2048}$, object classification confidence scores ${{\emph {\textbf{s}}}_i}\in \mathbb R^{O}$, and the spatial feature ${{\emph {\textbf{b}}}_i}\in \mathbb R^{4}$. Then, the concatenated feature is projected into a 512-dimensional subspace and denoted as ${{\emph {\textbf{x}}}_i}$. Besides, we further extract features from the union box of one pair of proposal $i$ and $j$, denoted as ${{\emph {\textbf{u}}}_{ij}}\in \mathbb R^{2048}$. To better capture properties of scene graph, we make contributions from three perspectives. First, a direction-aware message passing (DMP) module is introduced in Section \ref{setion1}. Second, a node priority sensitive loss (NPS-loss) is introduced in Section \ref{setion2}. Third, an adaptive reasoning module (ARM) is designed in Section \ref{setion3}.
\subsection{Direction-aware Message Passing}\label{setion1}
\begin{figure*}[tbp]
\setlength\abovecaptionskip{-1.8\baselineskip}
\setlength\belowcaptionskip{2pt}
\begin{center}
\includegraphics[scale=0.956]{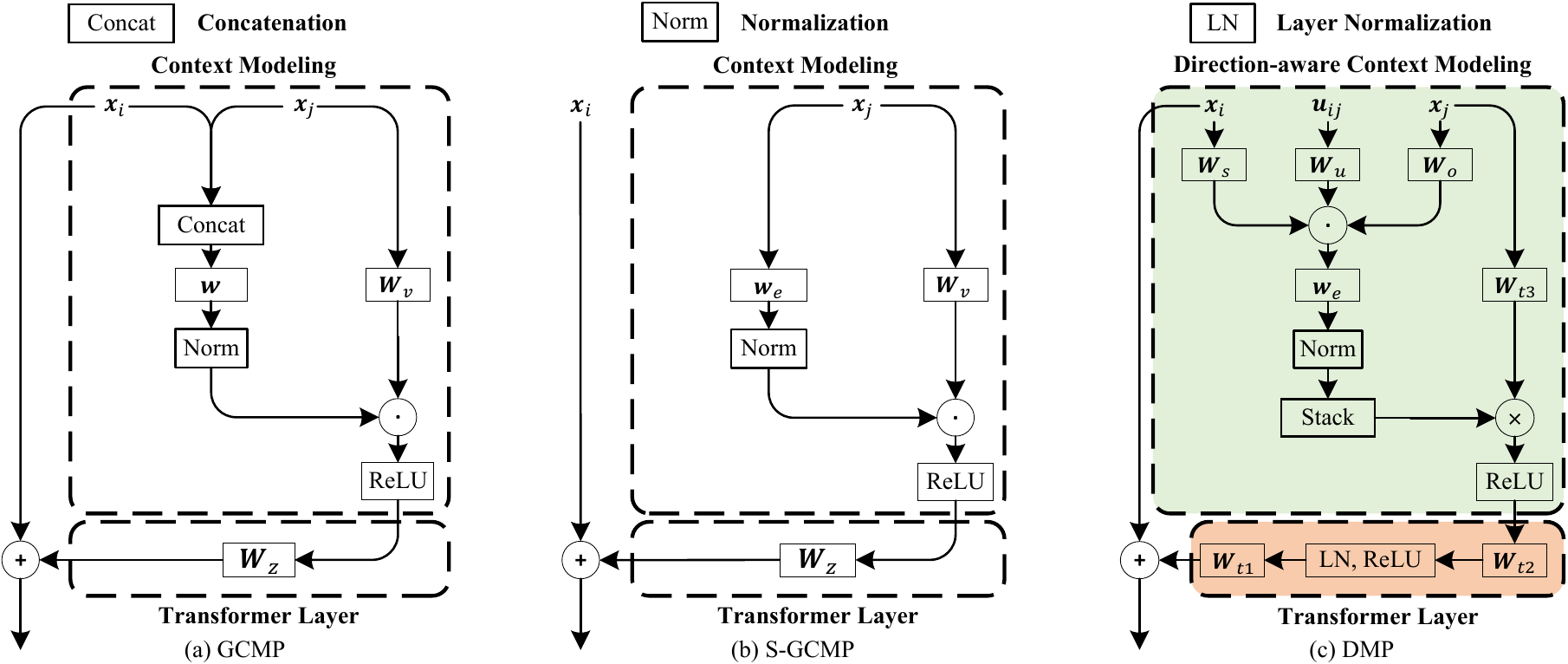}
\topcaption{Architecture of the three MP modules in Section \ref{setion1}. $\odot, \oplus,  \otimes, $ represent Hadamard product, element-wise addition, and Kronecker product, respectively.}
\label{archi}
\end{center}
\end{figure*}
The message passing (MP) module takes a node features ${{\emph {\textbf{x}}}_i}$ as input. Its output for the $i$-th node is denoted as ${{\emph {\textbf{z}}}_i}$, and the neighborhood of this node is represented as $\mathcal N_i$. For all MP modules in this section, $\mathcal N_i$ includes all nodes but the $i$-th node itself. Following the definition in graph attention network \cite{gat}, given two nodes $i$ and $j$, we represent the direction of $i \rightarrow j$ as forward and $i \leftarrow j$ as backward for the $i$-th node. In the following, we first review the design of the one representative MP module, which is denoted as Global Context MP (GCMP) in this paper. GCMP adopts the softmax function for normalization. Its structure is illustrated in Figure \ref{archi}(a) and can be formally expressed as
\begin{equation}\label{edge_f1}
{{\emph {\textbf{z}}}_i} = {{\emph {\textbf{x}}}_i} + {{\emph {\textbf{W}}}_z}\sigma \!\left(\sum\nolimits_{j \in \mathcal N_i} {\frac{{\exp ({\emph {\textbf{w}}}^{T}\left[ {{{\emph {\textbf{x}}}_i},{{\emph {\textbf{x}}}_j}} \right])}}{{\sum\nolimits_{m \in \mathcal N_i} {\exp ({\emph {\textbf{w}}}^{T}\left[ {{{\emph {\textbf{x}}}_i},{{\emph {\textbf{x}}}_m}} \right])} }}}  {{{\emph {\textbf{W}}}_v}{{\emph {\textbf{x}}}_j}}\right),
\end{equation}where $\sigma$ represents the ReLU function. ${{\emph {\textbf{W}}}_v}$ and ${{\emph {\textbf{W}}}_z}\in \mathbb R^{512\times512}$ are linear transformation matrices. ${{\emph {\textbf{w}}}}\in \mathbb R^{1024}$ is a projection vector, and $\left[ , \right]$ represents the concatenation operation. For simplicity, we define
$c_{ij}={\frac{{\exp ({\emph {\textbf{w}}}^{T}\left[ {{{\emph {\textbf{x}}}_i},{{\emph {\textbf{x}}}_j}} \right])}}{{\sum\nolimits_{m \in \mathcal N_i} {\exp ({\emph {\textbf{w}}}^{T}\left[ {{{\emph {\textbf{x}}}_i},{{\emph {\textbf{x}}}_m}} \right])} }}}$
as the pairwise contextual coefficient between nodes $i$ and $j$ in the forward direction. However, it has been revealed that utilizing the concatenation operation in Equation (\ref{edge_f1}) may not obtain node-specific contextual information \cite{gcnet}. In fact, it is more likely that ${\emph {\textbf{x}}}_i$ in Equation (\ref{edge_f1}) is ignored by ${\emph {\textbf{w}}}$. Therefore, GCMP actually generates the same contextual information for all nodes.

Inspired by this observation, Equation (\ref{edge_f1}) can be simplified as follows \cite{gcnet}:
\begin{equation}\label{gc}
{{\emph {\textbf{z}}}_i} = {{\emph {\textbf{x}}}_i} + {{\emph {\textbf{W}}}_z}\sigma\!\left(\sum\nolimits_{j \in \mathcal N_i} {\frac{{\exp ({\emph {\textbf{w}}}_e^{T} {{{\emph {\textbf{x}}}_j}} )}}{{\sum\nolimits_{m \in \mathcal N_i} {\exp ({\emph {\textbf{w}}}_e^{T} {{{\emph {\textbf{x}}}_m}})} }}}  {{{\emph {\textbf{W}}}_v}{{\emph {\textbf{x}}}_j}}\right),
\end{equation}
where ${{\emph {\textbf{w}}}}_e\in \mathbb R^{512}$ is a projection vector. As depicted in Figure \ref{archi}(b), we denote this model as Simplified Global Context MP (S-GCMP) module. The above two MP modules may not be optimal for SGG because they ignore the edge direction information and cannot provide node-specific contextual information. Accordingly, we propose the DMP module to solve the above problems. As illustrated in Figure \ref{archi}(c), DMP consists of two main components: \textbf{direction-aware context modeling} and one \textbf{transformer layer}.

\textbf{Direction-aware Context Modeling:}
This component aims to learn node-specific context and guide message passing via the edge direction information. Inspired by the multi-modal low rank bilinear pooling method \cite{MLB}, we formulate the contextual coefficient $e_{ij}$ between two nodes $i$ and $j$  as follows:
\begin{equation}\label{edgeq}
{{e_{ij}}} = {{\emph {\textbf{w}}}_{e}^{T}\left( {{{\emph {\textbf{W}}}_s}{{\emph {\textbf{x}}}_i} \odot {{\emph {\textbf{W}}}_o}{{\emph {\textbf{x}}}_j} \odot {{\emph {\textbf{W}}}_u}{{\emph {\textbf{u}}}_{ij}}} \right)},
\end{equation}where $\odot$ represents Hadamard product. ${{\emph {\textbf{W}}}_s}$, ${{\emph {\textbf{W}}}_o}$, and ${{\emph {\textbf{W}}}_u} \in {\mathbb R}^{512\times 512}$ are projection matrices for fusion. Equation (\ref{edgeq}) can be considered as a tri-linear model based on Tucker decomposition \cite{mutan}.

Compared with the first two MP modules, Equation (\ref{edgeq}) has four advantages. First, it employs union box features to expand the receptive field in context modeling. Second, the tri-linear model is a more powerful way to model high-order interactions between three types of features. Third, since features for the two nodes and the union box are coupled together by Hadamard product in Equation (\ref{edgeq}), they jointly affect context modeling. In this way, we obtain node-specific contextual information. Fourth, Equation (\ref{edgeq}) specifies the position of subject and object; therefore, it considers the edge direction information of the edge.

However, the direction of the edge is unclear in the MP step of SGG, since the relationship between two nodes is still unknown. Therefore, we consider the contextual coefficient for both the forward and backward directions by stacking them as a two-element-vector ${\left[ {\alpha _{ij}} {\alpha _{ji}}\right] }^{T}$, where $\alpha_{ij}$ denotes the normalized contextual coefficient. Finally, the output of the first component of DMP for the $i$-th node can be denoted as
\begin{equation}\label{out}
\sum\nolimits_{j \in \mathcal N_i} {\left[ \begin{array}{l}
{\alpha _{ij}}\\
{\alpha _{ji}}
\end{array} \right] \otimes }{\emph {\textbf{W}}}_{t3}{{{\emph {\textbf{x}}}_j}},
\end{equation}
where $\otimes$ denotes Kronecker product. $\emph{\textbf{W}}_{t3} \in {\mathbb R}^{256 \times {512}}$ is a learnable projection matrix.

\textbf{Transformer Layer:}
The contextual information obtained above may contain redundant information. Inspired by \cite{gcnet}, we employ a transformer layer to refine the obtained contextual information. Specifically, it is consisted of two fully-connected layers with ReLU activation and layer normalization (LN) \cite{LN}. Finally, residual connection is applied to fuse the original feature and the contextual information. Our whole DMP module can be expressed as
\begin{equation}\label{all}
{{{\emph {\textbf{z}}}_i}} = {{\emph {\textbf{x}}}_i} + {{{\emph {\textbf{W}}}_{t1}}}{\sigma}\! \!\left( {\mathop{\rm LN}\!\left( \! {{{\emph {\textbf{W}}}_{t2}} \! \sum\nolimits_{j \in \mathcal N_i} {\left[ \begin{array}{l}
{\alpha _{ij}}\\
{\alpha _{ji}}
\end{array} \right] \otimes }{\emph {\textbf{W}}}_{t3}{{{\emph {\textbf{x}}}_j}} } \right)} \right),
\end{equation}
where ${{{\emph {\textbf{W}}}_{t1}}\in {\mathbb R}^{512 \times  128}} $  and ${{{\emph {\textbf{W}}}_{t2}}\in {\mathbb R}^{128\times 512}}$ denote linear transformation matrices.


\subsection{Node Priority Sensitive Loss}\label{setion2}
Existing works for SGG tend to utilize cross-entropy loss as objective function for object classification, which implicitly regards the priority of all nodes is equal for the scene graph. However, their priority varies according to the number of triplets they are involved. Recently, a local-sensitive loss has been proposed to address this problem in \cite{agent}. As the loss is non-differentiable, the authors in \cite{agent} adopt a two-stage training strategy, where the second stage is realized by a complicated policy gradient method \cite{evolved}.
\begin{figure}[]
\setlength\abovecaptionskip{-1.6\baselineskip}
\setlength\belowcaptionskip{2pt}
\begin{center}
\includegraphics[scale=0.32]{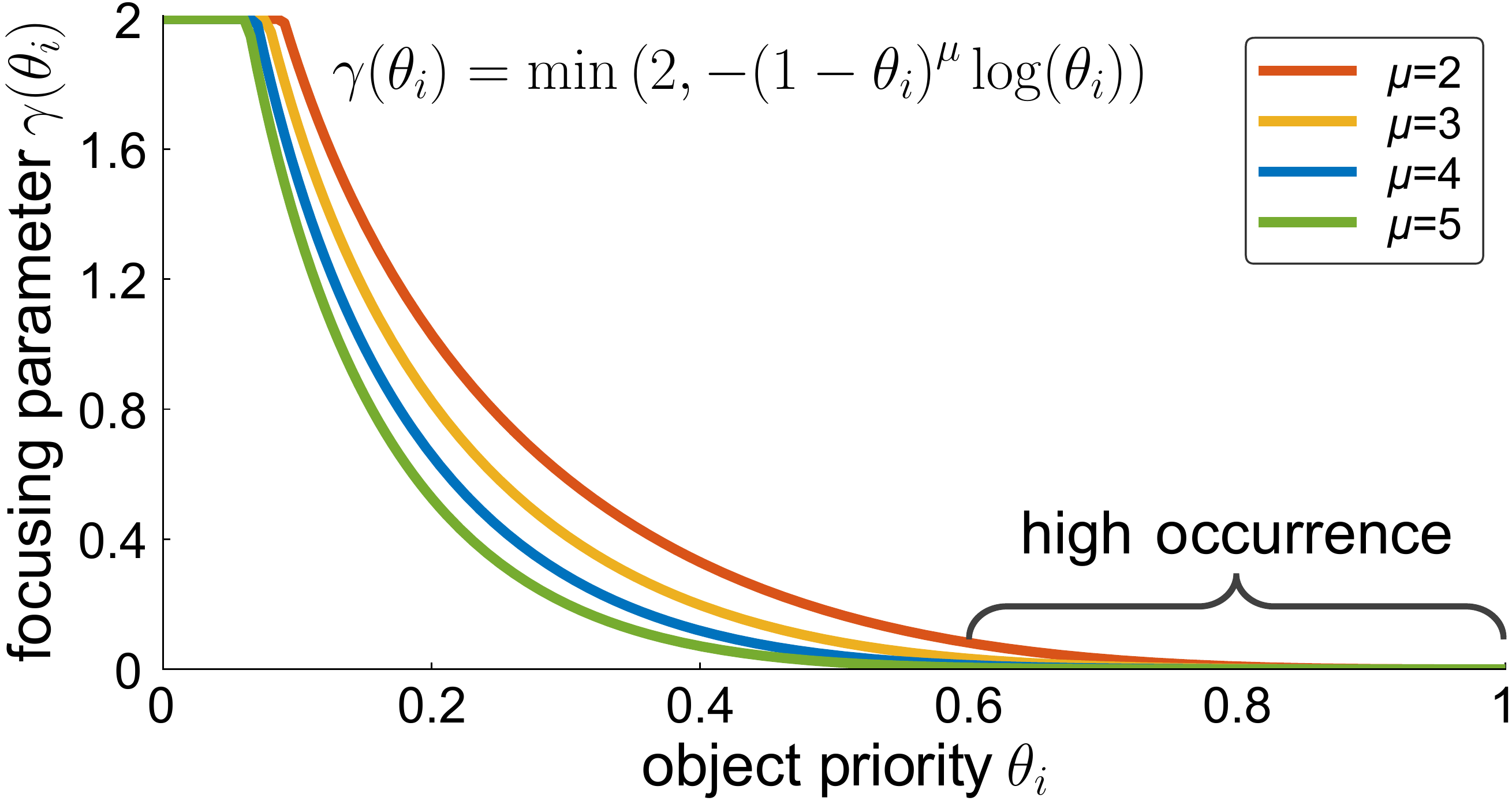}
\topcaption{Mapping function $\gamma(\theta_i)$ with different controlling factors $\mu$.}
\label{loss}
\end{center}
\end{figure}

To handle this problem, we propose a novel NPS-loss that not only captures the node priority in scene graph but also has the benefit of differentiable and convex formulation. NPS-loss is inspired by focal loss that reduces weights of well-classified objects using a focusing parameter, which is denoted as $\gamma$ in this paper. Compared with focal loss, NP-loss has the following key differences: (1) it is mainly used to solve the node-priority problem in SGG. In comparison, focal loss is designed to solve the class imbalance problem in object detection; (2) $\gamma$ is fixed in \cite{focalloss}. In NPS-loss, it depends on the node priority. Specifically, we first calculate the priority $\theta_i$ for the $i$-th node according to its contribution to the scene graph:
\begin{equation}\label{ratio}
 \theta_i = \frac{{ {{{ {t}}_{i}}} }}{\left\| T \right\|},
\end{equation}where $t_i$ denotes the number of triplets that include the $i$-th node and ${\left\| T \right\|}$ is the total number of triplets in one graph. Given $\theta_i$, one intuitive way to obtain the focusing parameter $\gamma$ is a linear transformation, \textit {e.g.}, $\gamma(\theta_i)=-2\theta_i+2$. However, this transformation exaggerates the difference between nodes of high-priority and middle-level priority, and narrows the difference between nodes of middle-level priority and low-priority. To solve this problem, we design a nonlinear mapping function that transforms $\theta_i$ to $\gamma$:
\begin{equation}\label{fl_re}
{\gamma}(\theta_i) = \min \left( {2, - {{(1 - {\theta _i})}^\mu }\log ({\theta _i})} \right),
\end{equation}where $\mu$ denotes a controlling factor, which controls the influence of $\theta_i$ to the value of $\gamma$. As depicted in Figure \ref{loss}, curve for the mapping function changes quickly for nodes with low priority, and slowly for nodes of high priority. Moreover, a larger $\mu$ leads to more nodes to be highlighted during training. Finally, we obtain the NPS-loss that guides the training process according to node priority:
\begin{equation}\label{nps}
{\mathcal L} _{nps}(p_i) = -(1-p_i)^{\gamma(\theta_i)}\log(p_i),
\end{equation}where $p_i$ denotes the object classification score on the ground-truth object class for the $i$-th node.

%
%
%
%
%

\subsection{Adaptive Reasoning Module}\label{setion3}
\begin{table*}[t]
\setlength\abovecaptionskip{-1\baselineskip}
\setlength\belowcaptionskip{2pt}
\setlength{\tabcolsep}{1.06mm}{\begin{tabular}{l|l|clclc|clclc|clclc|cl}
\hline
 &  & \multicolumn{5}{c|}{SGDET} & \multicolumn{5}{c|}{SGCLS} & \multicolumn{5}{c|}{PREDCLS} & \multicolumn{2}{c}{} \\
 & Model & R@20 & \multicolumn{1}{c}{} & R@50 & \multicolumn{1}{c}{} & R@100 & R@20 & \multicolumn{1}{c}{} & R@50 & \multicolumn{1}{c}{} & R@100 & R@20 & \multicolumn{1}{c}{} & R@50 & \multicolumn{1}{c}{} & R@100 & \multicolumn{2}{c}{Mean} \\ \hline\hline
 & GPI$^\diamond$ \cite{gpi} & \multicolumn{1}{c}{-} &  & \multicolumn{1}{c}{-} &  & \multicolumn{1}{c|}{-} & \multicolumn{1}{c}{-} &  & \multicolumn{1}{c}{36.5} &  & \multicolumn{1}{c|}{38.8} & \multicolumn{1}{c}{-} &  & \multicolumn{1}{c}{65.1} &  & \multicolumn{1}{c|}{66.9} & \multicolumn{2}{c}{-} \\
 Two-Stage & VCTREE-HL$^\diamond$ \cite{vctree} & \multicolumn{1}{c}{22.0} &  & \multicolumn{1}{c}{27.9} &  & \multicolumn{1}{c|}{31.3} & \multicolumn{1}{c}{35.2} &  & \multicolumn{1}{c}{38.1} &  & \multicolumn{1}{c|}{38.8} & \multicolumn{1}{c}{60.1} &  & \multicolumn{1}{c}{66.4} &  & \multicolumn{1}{c|}{68.1} & \multicolumn{2}{c}{45.1} \\
  & CMAT$^\diamond$ \cite{agent} & \multicolumn{1}{c}{22.1} &  & \multicolumn{1}{c}{27.9} &  & \multicolumn{1}{c|}{31.2} & \multicolumn{1}{c}{35.9} &  & \multicolumn{1}{c}{39.0} &  & \multicolumn{1}{c|}{39.8} & \multicolumn{1}{c}{60.2} &  & \multicolumn{1}{c}{66.4} &  & \multicolumn{1}{c|}{68.1} & \multicolumn{2}{c}{45.4} \\ \hline
 & IMP$^\diamond$ \cite{imp} & 14.6 &  & 20.7 &  & 24.5 & 31.7 &  & 34.6 &  & 35.4 & 52.7 &  & 59.3 &  & 61.3 & \multicolumn{2}{c}{39.3} \\
 & FREQ$^\diamond$\cite{neural-motif} & 20.1 &  & 26.2 &  & 30.1 & 29.3 &  & 32.3 &  & 32.9 & 53.6 &  & 60.6 &  & 62.2 & \multicolumn{2}{c}{40.7} \\
 & MOTIFS$^\diamond$ \cite{neural-motif} & 21.4 &  & 27.2 &  & 30.3 & 32.9 &  & 35.8 &  & 36.5 & 58.5 &  & 65.2 &  & 67.1 & \multicolumn{2}{c}{43.7} \\
One-Stage& Graph-RCNN \cite{graphrcnn} & - &  & 11.4 &  & 13.7 & - &  & 29.6 &  & 31.6 & - &  & 54.2 &  & 59.1 & \multicolumn{2}{c}{33.2} \\
 & KERN$^\diamond$  \cite{kern} & - &  & 27.1 &  & 29.8 & - &  & 36.7 &  & 37.4 & - &  & 65.8 &  & 67.6 & \multicolumn{2}{c}{44.1} \\
 & VCTREE-SL$^\diamond$ \cite{vctree} & 21.7 &  & 27.7 &  & 31.1 & 35.0 &  & 37.9 &  & 38.6 & 59.8 &  & 66.2 &  & 67.9 & \multicolumn{2}{c}{44.9} \\
 & CMAT-XE$^\diamond$ \cite{agent} & - &  & - &  & - & 34.0 &  & 36.9 &  & 37.6 & - &  & - &  & - & \multicolumn{2}{c}{-} \\
 & RelDN$^\ddagger $ \cite{contrastive} & 21.1 &  & 28.3 &  & 32.7 & 36.1 &  & 36.8 &  & 36.8 & 66.9 &  & 68.4 &  & 68.4 & \multicolumn{2}{c}{45.2} \\
 & \bf GPS-Net$^\diamond$ & \multicolumn{1}{c}{\bf22.6} &  & \multicolumn{1}{c}{\bf28.4} &  & \multicolumn{1}{c|}{\bf31.7} & \multicolumn{1}{c}{\bf36.1} &  & \multicolumn{1}{c}{\bf39.2} &  & \multicolumn{1}{c|}{\bf40.1} & \multicolumn{1}{c}{\bf60.7} &  & \multicolumn{1}{c}{\bf66.9} &  & \multicolumn{1}{c|}{\bf68.8} & \multicolumn{2}{c}{\bf45.9} \\
 & \bf GPS-Net$^\ddagger$ & \multicolumn{1}{c}{\bf22.3} &  & \multicolumn{1}{c}{\bf28.9} &  & \multicolumn{1}{c|}{\bf33.2} & \multicolumn{1}{c}{\bf41.8} &  & \multicolumn{1}{c}{\bf42.3} &  & \multicolumn{1}{c|}{\bf42.3} & \multicolumn{1}{c}{\bf67.6} &  & \multicolumn{1}{c}{\bf69.7} &  & \multicolumn{1}{c|}{\bf69.7} & \multicolumn{2}{c}{\bf47.7} \\ \hline
\end{tabular}}
\topcaption{Comparisons with state-of-the-arts on VG. Since some works do not evaluate on R@20, we compute the mean on all tasks over R@50 and R@100. $^\diamond$ and $^\ddagger$ denote the methods using the same Faster-RCNN detector and evaluation metric as \cite{neural-motif} and \cite{contrastive}, respectively.}
\label{table1}
\end{table*}
After obtaining the refined node features by DMP and the object classification scores by NPS-loss, we further propose an adaptive reasoning module (ARM) for relationship classification. Specifically, ARM provides prior for classification by two steps: frequency softening and bias adaptation for each triplet. In what follows, we introduce the two steps in detail.

\textbf{Frequency Softening:}
Inspired by the frequency baseline introduced in \cite{neural-motif}, we employ the frequency of relationships as prior to promote the performance of relationship classification. However, the original method in \cite{neural-motif} suffers from the long-tailed distribution problem of relationships. Therefore, it may fail to recognize relationships of low frequency. To handle this problem, we first adopt a log-softmax function to soften the original frequency distribution of relationships as follows:
\begin{equation}\label{log_st}
{{\tilde {\emph {\textbf{p}}}}^{i \to j}}  = {\log \rm softmax}\left( {{{\emph {\textbf{p}}}}^{i \to j}} \right),
\end{equation}
where ${{{\emph {\textbf{p}}}}^{i \to j}} \in \mathbb{R}^{{R}}$ denotes the original frequency distribution vector between the $i$-th and the $j$-th nodes. The same as \cite{neural-motif}, this vector is determined by the object class of the two nodes. ${{\tilde {\emph {\textbf{p}}}}^{i \to j}}$ is the normalized vector of ${{{\emph {\textbf{p}}}}^{i \to j}}$.

\textbf{Bias Adaptation:}
To enable the frequency prior adjustable for each node pair, we further propose an adpative attention mechanism to modify the prior according to the visual appearance of the node pair. Specifically, a sigmoid function is applied to obtain attention on the frequency prior: ${\emph {\textbf{d}}}  = {\rm sigmoid} \left( {{{\emph {\textbf{W}}}_p}{{\emph {\textbf{u}}}_{ij}} } \right)$, where ${{\emph {\textbf{W}}}_p}\in \mathbb R^{{R}\times2048}$ is transformation matrix. Then, the classification score vector of relationships can be obtained as follows:
\begin{equation}\label{sd}
  {{\emph {\textbf{p}}}_{ij}} = {\rm softmax} \left( {{{\emph {\textbf{W}}}_r}( {{{\emph {\textbf{z}}}_i}*{{\emph {\textbf{z}}}_j}} * {{{\emph {\textbf{u}}}_{ij}}} )+ {\emph {\textbf{d}}} \odot  {{\tilde {\emph {\textbf{p}}}}^{i \to j}}} \right),
\end{equation}where ${{\emph {\textbf{W}}}_r}\in \mathbb R^{{ R}\times1024}$ denotes the classifier, and ${\emph {\textbf{d}}} \odot  {{\tilde {\emph {\textbf{p}}}}^{i \to j}}$ is the bias. $*$ represents a fusion function defined in \cite{fusion}: \begin{table}[h]
\setlength\abovecaptionskip{-1\baselineskip}
\setlength\belowcaptionskip{11pt}
\setlength{\tabcolsep}{1.5mm}
\begin{tabular}{l|c|c|c}
\hline
 & SGDET & SGCLS & PREDCLS \\
Model & mR@100 & mR@100 & mR@100 \\ \hline\hline
IMP$^\diamond$ \cite{imp} & 4.8 & 6.0 & 10.5 \\
FREQ$^\diamond$ \cite{neural-motif} & 7.1 & 8.5 & 16.0 \\
MOTIFS$^\diamond$ \cite{neural-motif} & 6.6 & 8.2 & 15.3 \\
KERN$^\diamond$  \cite{kern} & 7.3 & 10.0 & 19.2 \\
VCTREE-HL$^\diamond$ \cite{vctree} & 8.0 & 10.8 & 19.4 \\ \hline
\bf GPS-Net$^\diamond$ & \bf 9.8 & \bf 12.6 & \bf 22.8 \\ \hline
\end{tabular}
\par\nointerlineskip\vspace{-3.0mm}
\topcaption{Comparison on the mR@100 metric between various methods across all the 50 relationship categories.}
\label{table2}
\end{table} ${\emph {\textbf{x}}} * {\emph {\textbf{y}}} = {\mathop{\rm ReLU}\nolimits} \left( {{{\emph {\textbf{W}}}_x}{\emph {\textbf{x}}} + {{\emph {\textbf{W}}}_y}{\emph {\textbf{y}}}} \right) - \left( {{{\emph {\textbf{W}}}_x}{\emph {\textbf{x}}} - {{\emph {\textbf{W}}}_y}{\emph {\textbf{y}}}} \right) \odot \left( {{{\emph {\textbf{W}}}_x}{\emph {\textbf{x}}} - {{\emph {\textbf{W}}}_y}{\emph {\textbf{y}}}} \right)$, where ${{\emph {\textbf{W}}}_x}$ and ${{\emph {\textbf{W}}}_y}$ project ${{\emph {\textbf{x}}}},{{\emph {\textbf{y}}}}$ to 1024-dimensional space, respectively.

\textbf{Relationship Prediction:}
During testing, the category of relationship between $i$-th and $j$-th nodes is predicted by:
\begin{equation}\label{1}
r_{ij}={\arg{\max} _{r \in {\mathcal R}}}({{\emph {\textbf{p}}}_{ij}(r)}),
\end{equation}
where $\mathcal R$ represents the set of relationship categories.

\section{Experiments}
We present experimental results on three datasets: Visual Genome (VG) \cite{VGdataset}, OpenImages (OI) \cite{OI}, and Visual Relationship Detection (VRD) \cite{vrd}. We first report evaluation settings, followed by comparisons with state-of-the-art methods and the ablation studies. Besides, qualitative comparisons between GPS-Net and other approaches are provided in the supplementary file.
\subsection{Evaluation Settings}
 \begin{table*}[t]
\setlength\abovecaptionskip{0.2\baselineskip}
\setlength\belowcaptionskip{-13pt}
\setlength{\tabcolsep}{0.65mm}{
\begin{tabular}{l|cccc|ccccccccc}
\hline
 &  & \multicolumn{1}{l}{} &  & \multicolumn{1}{l|}{} & \multicolumn{9}{c}{${\rm {AP}}_{rel}$ per class} \\
Model & R@50 & ${\rm {wmAP}}_{rel}$ & ${\rm {wmAP}}_{phr}$ & ${\rm {score}}_{wtd}$ & at & on & holds & plays & interacts with & wears & hits & inside of & under  \\ \hline \hline
RelDN, $L_0$\cite{contrastive} & 74.67 & 34.63 & 37.89 & 43.94 & 32.40 & 36.51 & 41.84 & 36.04 & 40.43 & 5.70 & 55.40 & 44.17 & 25.00  \\
RelDN\cite{contrastive} & 74.94 & 35.54 & 38.52 & 44.61 & 32.90 & 37.00 & 43.09 & 41.04 & 44.16 & 7.83 & 51.04& 44.72 & 50.00  \\ \hline
{\textbf {GPS-Net}} & \bf 77.27 & \bf 38.78 &  \bf 40.15 & \bf 47.03 & \bf 35.10 & \bf 38.90 & \bf 51.47 &\bf  45.66 & \bf 44.58 & \bf 32.35 & \bf 71.71 & \bf 47.21 & \bf 57.28  \\ \hline
\end{tabular}}
\caption{Comparisons with state-of-the-arts on OI. We adopt the same evaluation metrics as \cite{contrastive}}
\label{table6}
\end{table*}
\textbf{Visual Genome:}
We use the same data and evaluation metrics that have been widely adopted in recent works \cite{graphrcnn, attentive, MSDN, gpi, factorize, agent}. Specifically, the most frequent 150 object categories and 50 relationship categories are utilized for evaluation. After preprocessing, the scene graph for each image consists of 11.6 objects and 6.2 relationships on average. The data is divided into one training set and one testing set. The training set includes 70$\%$ images, with 5K images as a validation subset. The testing set is composed of the remaining 30$\%$ images. In the interests of fair comparisons, we also adopt Faster R-CNN \cite{fasterrcnn} \begin{figure}[ht]
\setlength\abovecaptionskip{0.4\baselineskip}
\setlength\belowcaptionskip{-26pt}
\begin{center}
\includegraphics[scale=0.71]{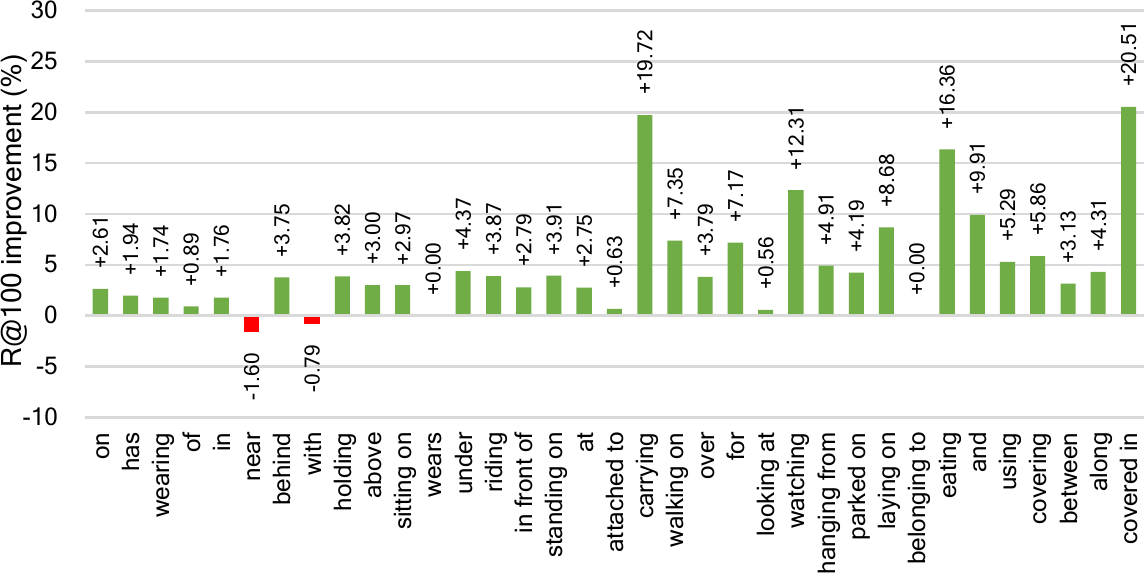}
\par\nointerlineskip\vspace{-2mm}
\caption{The R@100 improvement in PREDCLS of GPS-Net compared with the VCTREE \cite{vctree}. The Top-35 categories of relationship are selected according to their occurrence frequency.}
\label{meanrecall12}
\end{center}
\end{figure}with VGG-16 backbone to obtain the location and features of object proposals. Moreover, since SGG performance highly depends on the pre-trained object detector, we utilize the same set of hyper-parameters as \cite{neural-motif} and \cite{contrastive} respectively. We follow three conventional protocols for evaluation: (1) Scene Graph Detection (SGDET): given an image, detect object bounding boxes and their categories, and predict their pair-wise relationships; (2) Scene Graph Classification (SGCLS): given ground-truth object bounding boxes, predict the object categories and their pair-wise relationships; (3) Predicate Classification (PREDCLS): given the object categories and their bounding boxes, predict their pair-wise relationships only. All algorithms are evaluated by Recall@$K$ metrics, where $K$=20, 50, and 100, respectively. Considering that the distribution of relationships is highly imbalanced in VG, we further utilize mean recall@K (mR@K) to evaluate the performance of each relationship \cite{vctree, kern}.

\textbf{OpenImages:}
The training and testing sets contain 53,953 images and 3,234 images respectively. We utilize Faster R-CNN associated with the pre-trained ResNeXt-101-FPN \cite{contrastive} as the backbone. We also follow the same data processing and evaluation metrics as in \cite{contrastive}. More specifically,
\begin{table}
\setlength\abovecaptionskip{0.2\baselineskip}
\setlength\belowcaptionskip{-15pt}
\setlength{\tabcolsep}{0.18mm}{
\begin{tabular}{lccccccc}
\hline
\multicolumn{1}{l|}{} & \multicolumn{1}{c|}{Pre.} & \multicolumn{3}{c|}{Rel.} & \multicolumn{3}{c}{Phr.} \\
\multicolumn{1}{l|}{Model} & \multicolumn{1}{c|}{R@50} & R@50 & \multicolumn{1}{l}{} & \multicolumn{1}{c|}{R@100} & R@50 & \multicolumn{1}{l}{} & R@100 \\ \hline\hline
\multicolumn{1}{l|}{VTransE \cite{Vtranse}} & \multicolumn{1}{c|}{44.8} & 19.4 &  & \multicolumn{1}{c|}{22.4} & 14.1 &  & 15.2 \\
\multicolumn{1}{l|}{ViP-CNN  \cite{ViP-CNN}} & \multicolumn{1}{c|}{-} & 17.3 &  & \multicolumn{1}{c|}{20.0} & 22.8 &  & 27.9 \\
\multicolumn{1}{l|}{VRL \cite{VRL}} & \multicolumn{1}{c|}{-} & 18.2 &  & \multicolumn{1}{c|}{20.8} & 21.4 &  & 22.6 \\
\multicolumn{1}{l|}{{KL distilation \cite{KL}}} & \multicolumn{1}{c|}{55.2} & 19.2 &  & \multicolumn{1}{c|}{21.3} & 23.1 &  & 24.0 \\
\multicolumn{1}{l|}{MF-URLN \cite{zhan}} & \multicolumn{1}{c|}{58.2} & 23.9 &  & \multicolumn{1}{c|}{26.8} & 31.5 &  & 36.1 \\\hline
\multicolumn{1}{l|}{Zoom-Net$^* $\cite{ZOOM-NET}} & \multicolumn{1}{c|}{50.7} & 18.9 &  & \multicolumn{1}{c|}{21.4} & 24.8 &  & 28.1 \\
\multicolumn{1}{l|}{CAI + SCA-M$^*$\cite{ZOOM-NET}} & \multicolumn{1}{c|}{56.0} & 19.5 &  & \multicolumn{1}{c|}{22.4} & 25.2 &  & 28.9 \\
\multicolumn{1}{l|}{\textbf{GPS-Net}$^*$ (ImageNet)} & \multicolumn{1}{c|}{\bf 58.7} & \bf 21.5 &  &  \multicolumn{1}{c|}{\bf 24.3} & \textbf{28.9} &  & \textbf{34.0} \\\hline
\multicolumn{1}{l|}{RelDN$^\dag $  \cite{contrastive}} & \multicolumn{1}{c|}{-} & 25.3 &  & \multicolumn{1}{c|}{28.6} & 31.3 &  & 36.4 \\
\multicolumn{1}{l|}{\textbf{GPS-Net}$^\dag $ (COCO)} & \multicolumn{1}{c|}{\bf 63.4} & \bf 27.8 &  &  \multicolumn{1}{c|}{\bf 31.7} & \textbf{33.8} &  & \textbf{39.2} \\ \hline
\end{tabular}}
\caption{Comparisons with state-of-the-arts on VRD ($-$ denotes unavailable). Pre., Phr., and Rel. represent predication detection, phrase detection, and relation detection, respectively.
$^\dag $ and $^*$ denote using the same object detector.
}
\label{table5}
\end{table}the results are evaluated by calculating Recall@50 (R@50), weighted mean AP of relationships $({\rm {wmAP}}_{rel})$, and weighted mean AP of phrase (${\rm {wmAP}}_{phr}$). The final score is given by score$_{wtd}=0.2\times R@50+0.4\times {\rm {wmAP}}_{rel}+0.4\times {\rm {wmAP}}_{phr}$. Note that the ${\rm {wmAP}}_{rel}$ evaluates the AP of the predicted triplet where both the subject and object boxes have an IoU of at least 0.5 with ground truth. The ${\rm {wmAP}}_{phr}$ is similar, but utilized for the union area of the subject and object boxes.

\textbf{Visual Relationship Detection:}
We apply the same object detectors as in \cite{contrastive}. More specifically, two VGG16-based backbones are provided, which were trained on ImgaeNet and COCO, respectively. The evaluation metric is the same as in \cite{vrd}, which reports R@50 and R@100 for relationship, predicate, and phrase detection.

\subsection{Implementation Details}
To ensure compatibility with the architectures of previous state-of-the-art methods, we utilize ResNeXt-101-FPN as our OpenImages backbone on OI and VGG-16 on VG and VRD. During training, we freeze the layers before the ROIAlign layer and optimize the model jointly considering the object and relationship classification losses. Our model is optimized by SGD with momentum, with the initial learning rate and batch size set to 10$^{-3}$ and 6 respectively. For the SGDET task, we follow \cite{neural-motif} that we only predict the relationship between proposal pairs with overlapped bounding boxes. Besides, the top-64 object proposals in each image are selected after per-class non-maximal suppression (NMS) with an IoU of 0.3. Moreover, the ratio between pairs without any relationship (background pairs) and those with relationship during training is sampled to 3:1.
\subsection{Comparisons with State-of-the-Art Methods}

\begin{table*}
\setlength\abovecaptionskip{-0.7\baselineskip}
\setlength\belowcaptionskip{-9pt}
\setlength{\tabcolsep}{1.5mm}\begin{tabular}{llclclclclclcclclccllll}
\hline
\multicolumn{2}{l|}{} &\multicolumn{6}{c|}{Module} & \multicolumn{5}{c|}{SGDET} & \multicolumn{5}{c|}{SGCLS} & \multicolumn{5}{c}{PREDCLS} \\
\multicolumn{2}{c|}{Exp} &\multicolumn{2}{c}{DMP} & \multicolumn{2}{c}{NPS} & \multicolumn{2}{c|}{ARM} & R@20 &  & R@50 &  & \multicolumn{1}{c|}{R@100} & R@20 &  & R@50 &  & \multicolumn{1}{c|}{R@100} & \multicolumn{1}{l}{R@20} &  & \multicolumn{1}{l}{R@50} &  & \multicolumn{1}{l}{R@100} \\ \hline\hline
\multicolumn{2}{c|}{1} &\multicolumn{2}{c}{} & \multicolumn{2}{c}{} & \multicolumn{2}{l|}{} &21.1  &  &26.3  &  & \multicolumn{1}{c|}{29.4} &32.7  &  &35.4  &  & \multicolumn{1}{c|}{36.3} &\multicolumn{1}{c}{{58.8}}  &  &\multicolumn{1}{c}{{65.6}}  &  &\multicolumn{1}{c}{{67.3}}\\
\multicolumn{2}{c|}{2} &\multicolumn{2}{c}{\checkmark} & \multicolumn{2}{l}{} & \multicolumn{2}{l|}{} &22.3
  &  &28.1  &  &   \multicolumn{1}{c|}{31.4} &35.2  &  &38.3 &  & \multicolumn{1}{c|}{39.3} &\multicolumn{1}{c}{{59.6}}  &  &\multicolumn{1}{c}{{66.1}}  &  &\multicolumn{1}{c}{{67.9}}  \\

\multicolumn{2}{c|}{3} &\multicolumn{2}{c}{} & \multicolumn{2}{c}{\checkmark} & \multicolumn{2}{l|}{} &21.5  &  &26.6  &  & \multicolumn{1}{c|}{29.8} &33.2  &  &36.3  &  & \multicolumn{1}{c|}{37.1} &\multicolumn{1}{c}{{59.1}}  &  &\multicolumn{1}{c}{{65.9}}  &  &\multicolumn{1}{c}{{67.7}}  \\

\multicolumn{2}{c|}{4} &\multicolumn{2}{l}{} & \multicolumn{2}{l}{} & \multicolumn{2}{c|}{\checkmark} &21.3  &  &26.5  &  & \multicolumn{1}{c|}{29.6} &32.9  &  &35.8  &  & \multicolumn{1}{c|}{36.8} &\multicolumn{1}{c}{{60.5}}  &  &\multicolumn{1}{c}{{66.7}}  &  &\multicolumn{1}{c}{{68.5}}  \\

%
%
%
%
%

\multicolumn{2}{c|}{5} &\multicolumn{2}{c}{\checkmark} & \multicolumn{2}{c}{\checkmark} & \multicolumn{2}{c|}{\checkmark} &\bf{22.6}  &  &\bf{28.4}  &  & \multicolumn{1}{c|}{\bf{31.7}} &\bf{36.1}  &  &\bf{39.2}  &  & \multicolumn{1}{c|}{\bf{40.1}} &\multicolumn{1}{c}{\bf{60.7}}   &  &\multicolumn{1}{c}{\bf{66.9}}   &  &\multicolumn{1}{c}{\bf{68.8}}  \\\hline
\label{table33}
\end{tabular}
{\topcaption{Ablation studies on the proposed methods. We consistently use the same backbone as \cite{neural-motif}. }
\label{tb3}}
\end{table*}
\begin{table*}
\setlength\abovecaptionskip{-1.2\baselineskip}
\setlength\belowcaptionskip{7pt}
\setlength{\tabcolsep}{0.32mm}\begin{tabular}{ccc|clclcl}
\hline
\multicolumn{3}{c|}{} & \multicolumn{2}{c}{} & \multicolumn{2}{c}{ w. stack} & \multicolumn{2}{c}{w.o. stack} \\ \hline\hline
 &  &  & \multicolumn{2}{c}{R@20} & \multicolumn{2}{c}{35.7} & \multicolumn{2}{c}{\bf36.1} \\
\multicolumn{3}{c|}{SGCLS} & \multicolumn{2}{c}{R@50} & \multicolumn{2}{c}{38.8} & \multicolumn{2}{c}{\bf39.2} \\
\multicolumn{1}{l}{} &  &  & \multicolumn{2}{c}{R@100} & \multicolumn{2}{c}{39.6} & \multicolumn{2}{c}{\bf40.1} \\ \hline
\multicolumn{1}{l}{} &  &  & \multicolumn{2}{c}{R@20} & \multicolumn{2}{c}{22.4} & \multicolumn{2}{c}{\bf22.6} \\
\multicolumn{3}{c|}{SGDET} & \multicolumn{2}{c}{R@50} & \multicolumn{2}{c}{28.3} & \multicolumn{2}{c}{\bf28.4} \\
\multicolumn{1}{l}{} & \multicolumn{1}{l}{} & \multicolumn{1}{l|}{} & \multicolumn{2}{c}{R@100} & \multicolumn{2}{c}{31.5} & \multicolumn{2}{c}{\bf31.7} \\ \hline\\
\end{tabular}
                \hfill
\setlength{\tabcolsep}{0.52mm}\begin{tabular}{ccc|clclclcl}
\hline
\multicolumn{3}{c|}{} & \multicolumn{2}{c}{} & \multicolumn{2}{c}{GCMP} & \multicolumn{2}{c}{S-GCMP} & \multicolumn{2}{l}{DMP} \\ \hline\hline
 &  &  & \multicolumn{2}{c}{R@20} & \multicolumn{2}{c}{34.3} & \multicolumn{2}{c}{34.8} & \multicolumn{2}{c}{\bf36.1} \\
\multicolumn{3}{c|}{SGCLS} & \multicolumn{2}{c}{R@50} & \multicolumn{2}{c}{37.2} & \multicolumn{2}{c}{37.7} & \multicolumn{2}{c}{\bf39.2} \\
\multicolumn{1}{l}{} &  &  & \multicolumn{2}{c}{R@100} & \multicolumn{2}{c}{37.9} & \multicolumn{2}{c}{38.4} & \multicolumn{2}{c}{\bf40.1} \\ \hline
\multicolumn{1}{l}{} &  &  & \multicolumn{2}{c}{R@20} & \multicolumn{2}{c}{21.7} & \multicolumn{2}{c}{22.1} & \multicolumn{2}{c}{\bf22.6} \\
\multicolumn{3}{c|}{SGDET} & \multicolumn{2}{c}{R@50} & \multicolumn{2}{c}{27.5} & \multicolumn{2}{c}{28.0} & \multicolumn{2}{c}{\bf28.4} \\
\multicolumn{1}{l}{} & \multicolumn{1}{l}{} & \multicolumn{1}{l|}{} & \multicolumn{2}{c}{R@100} & \multicolumn{2}{c}{30.8} & \multicolumn{2}{c}{31.2} & \multicolumn{2}{c}{\bf31.7} \\ \hline \\
\end{tabular}
\hfill
\setlength{\tabcolsep}{0.52mm}\begin{tabular}{ccc|clclclclcl}
\hline
\multicolumn{3}{c|}{} & \multicolumn{2}{c}{} & \multicolumn{2}{c}{Focal} & \multicolumn{2}{c}{$\mu =3$} & \multicolumn{2}{c}{$\mu =4$} & \multicolumn{2}{c}{$\mu =5$} \\ \hline\hline
 &  &  & \multicolumn{2}{c}{R@20} & \multicolumn{2}{c}{35.8} & \multicolumn{2}{c}{36.0} & \multicolumn{2}{c}{\bf36.1} & \multicolumn{2}{c}{35.8} \\
\multicolumn{3}{c|}{SGCLS} & \multicolumn{2}{c}{R@50} & \multicolumn{2}{c}{39.0} & \multicolumn{2}{c}{38.9} & \multicolumn{2}{c}{\bf39.2} & \multicolumn{2}{c}{39.1} \\
\multicolumn{1}{l}{} &  &  & \multicolumn{2}{c}{R@100} & \multicolumn{2}{c}{39.8} & \multicolumn{2}{c}{39.9} & \multicolumn{2}{c}{\bf40.1} & \multicolumn{2}{c}{39.9} \\ \hline
\multicolumn{1}{l}{} &  &  & \multicolumn{2}{c}{R@20} & \multicolumn{2}{c}{22.4} & \multicolumn{2}{c}{22.5} & \multicolumn{2}{c}{\bf22.6} & \multicolumn{2}{c}{22.5} \\
\multicolumn{3}{c|}{SGDET} & \multicolumn{2}{c}{R@50} & \multicolumn{2}{c}{28.2} & \multicolumn{2}{c}{28.2} & \multicolumn{2}{c}{\bf28.4} & \multicolumn{2}{c}{28.3} \\
\multicolumn{1}{l}{} & \multicolumn{1}{l}{} & \multicolumn{1}{l|}{} & \multicolumn{2}{c}{R@100} & \multicolumn{2}{c}{31.5} & \multicolumn{2}{c}{31.6} & \multicolumn{2}{c}{\bf31.7} & \multicolumn{2}{c}{31.6} \\ \hline\\
\end{tabular}
\par\nointerlineskip\vspace{-5.8mm}
{\topcaption{The {\bf{left sub-table}} shows the effectiveness of the stacking operation in DMP. The {\bf{middle sub-table}} compares the performance of the three MP modules in Section \ref{setion1} with the same transformer layer. The {\bf{right sub-table}} compares NPS-loss and the focal loss, and shows the influence of the controlling factor $\mu$. }\label{tb4}}
\end{table*}
\textbf{Visual Genome:} Table \ref{table1} shows that GPS-Net outperforms all state-of-the-arts methods on various metrics. Specifically, GPS-Net outperforms one very recent one-stage model, named KERN \cite{kern}, by 1.8$\%$ on average at R@50 and R@100 over the three protocols. In more detail, it outperforms KERN by 1.9$\%$, 2.7$\%$ and 1.2$\%$ at R@100 on SGDET, SGCLS, and PRECLS, respectively. Even when compared with the best two-stage model CMAT \cite{agent}, GPS-Net still demonstrates a performance improvement of 0.5$\%$ on average over the three protocols. Meanwhile, compared with the one-stage version of VCTREE \cite{vctree} and CMAT \cite{agent}, GPS-Net respectively achieves 1.5$\%$ and 2.5$\%$ performance gains on SGCLS at Recall@100. Another advantage of GPS-Net over VCTREE and CMAT is that GPS-Net is much more efficient, as the two methods adopt policy gradient for optimization, which is time-consuming \cite{evolved}. Moreover, when compare with RelDN using the same backbone, the performance gain by GPS-Net is even more dramatic, namely, 5.5$\%$ promotion on SGCLS at Recall@100 and 2.5$\%$ on average over three protocols.

Due to the class imbalance problem in VG, previous works usually achieve low performance for less frequent categories. Hence, we conduct an experiment utilizing the Mean Recall as evaluation metric \cite{kern, vctree}. As shown in Table \ref{table2} and Figure \ref{meanrecall12}, GPS-Net shows a large absolute gain for both the Mean Recall and Recall metrics, which indicates that GPS-Net has advantages in handling the class imbalance problem of SGG.

\textbf{OpenImages:}
We present results compared with RelDN  \cite{contrastive} in Table \ref{table6}. RelDN is an improved version of the model that won the Google OpenImages Visual Relationship Detection Challenge, with the same object detector, GPS-Net outperforms RelDN by 2.4$\%$ on the overall metric $score_{wtd}$. Moreover, despite of the severe class imbalance problem, GPS-Net still achieves outstanding performance in ${\rm {AP}}_{rel}$ for each category of relationships. The largest gap between GPS-Net and RelDN in ${\rm {AP}}_{rel}$ is 24.5$\%$  for \textit{wears} and 20.6$\%$ for \textit{hits}.

\textbf{Visual Relationship Detection:} Table \ref{table5} presents comparisons on VRD with state-of-the-art methods. To facilitate fair comparison, we adopt the two backbone models provided in RelDN \cite{contrastive} to train GPS-Net, respectively. It is shown that GPS-Net consistently achieves superior performance with both backbone models.
\subsection{Ablation Studies}
To prove the effectiveness of our proposed methods, we conduct four ablation studies. Results of the ablation studies are summarized in Table \ref{tb3} and Table \ref{tb4}, respectively.

\textbf{Effectiveness of the Proposed Modules.} We first perform an ablation study to validate the effectiveness of DMP, NPS-loss, and ARM. Results are summarized in Table 5. We add the above modules one by one to the baseline model. In Table \ref{tb3}, Exp 1 demotes our baseline that is based on the MOTIFNET-NOCONTEXT method \cite{neural-motif} with our feature construction strategy for relationship prediction. From Exp 2-5, we can clearly see that the performance improves consistently when all the modules are used together. This shows that each module plays a critical role in inferring object labels and their pair-wise relationships.

\textbf{Effectiveness of the Stacking Operation in DMP.} We conduct additional analysis on the stacking operation in DMP. The stacking operation accounts for the uncertainty in the edge direction information. As shown in the left sub-table of Table \ref{tb4}, the stacking operation consistently improves the performance of DMP over various metrics. Therefore, its effectiveness is justified.


\textbf{Comparisons between Three MP Modules.} We compare the performance of three MP modules in Section \ref{setion1}: GCMP, S-GCMP, and DMP. To facilitate fair comparison, we implement the same transformer layer as DMP to the other two modules. As shown in the middle sub-table in Table \ref{tb4}, the performance of DMP is much better than the other two modules. This is because DMP encodes the edge direction information and provides node-specific contextual information for each node involved in message passing.

\textbf{Design Choices in NPS-loss.} The value of the controlling factor $\mu$ determines the impact of node priority on object classification. As shown in the right sub-table of Table \ref{tb4}, we show the performance of NPS-loss with three different values of $\mu$. We also compare NPS-loss with the focal loss \cite{focalloss}. NPS-loss achieves the best performance when $\mu$ equals to 4. Moreover, NPS-loss outperforms the focal loss, justifying its effectiveness to solve the node priority problem for SGG.
\section{Conclusion}
In this paper, we devise GPS-Net to address the main challenges in SGG by capturing three key properties of scene graph. Specifically, (1) edge direction is encoded when calculating the node-specific contextual information via the DMP module; (2) the difference in node priority is characterized by a novel NPS-loss; and (3) the long-tailed distribution of relationships is alleviated by improving the usage of relationship frequency through ARM. Through extensive comparative experiments and ablation studies, we validate the effectiveness of GPS-Net on three datasets.
~\\

\noindent{\textbf {Acknowledgment}}.\quad Changxing Ding was supported in part by NSF of China under Grant 61702193, in part by the Science and Technology Program of Guangzhou under Grant 201804010272, in part by the Program for Guangdong Introducing Innovative and Entrepreneurial Teams under Grant 2017ZT07X183. Dacheng Tao was supported in part by ARC FL-170100117 and DP-180103424.


\newpage
\begin{appendices}
\section*{Appendix}

This supplementary document is organized as follows:
\begin{itemize}
  \item Section \ref{quall} shows more detailed comparison results under the {\bf Mean Recall} metric between GPS-Net and state-of-the-art methods on the VG database. Results are summarized in Table \ref{tba5}. Meanwhile, we also treat the number of relationship predictions per object pair ($k$) as a hyper-parameter on VRD, and report Recall with respect to different $k$ in Table \ref{tabvv}.
  \item Section \ref{qualiti} first provides the qualitative comparisons between GPS-Net and a strong baseline named MOTIFS \cite{neural-motif} under the SGDET protocol in Figure \ref{qualitative}. Then, attention maps of different MP modules are visualized in Figure \ref{atten}.
\end{itemize}

\section{Quantitative Analysis}\label{quall}
\subsection{Mean Recall for Scene Graph on VG}
As shown in Table \ref{tba5}, GPS-Net shows the best performance under all protocols. In particular, GPS-Net outperforms one very recent work named VCTREE-HL \cite{vctree} by 2.3$\%$ on average over the three protocols.
\begin{table*}[h]
\setlength\abovecaptionskip{-1\baselineskip}
\setlength\belowcaptionskip{5pt}
\setlength{\tabcolsep}{0.8mm}{
\begin{tabular}{llll|ccccc|ccccc|ccccc|c}
\hline
 &  &  &  & \multicolumn{5}{c|}{SGDET} & \multicolumn{5}{c|}{SGCLS} & \multicolumn{5}{c|}{PREDCLS} & \\
Model &  &  &  & mR@20 &  & mR@50 &  & mR@100 & mR@20 &  & mR@50 &  & mR@100 & mR@20 &  & mR@50 &  & mR@100 & Mean\\ \hline
IMP$^\diamond$ \cite{imp} &  &  &  & - &  & 3.8 &  & 4.8 & - &  & 5.8 &  & 6.0 & - &  & 9.8 &  & 10.5& 6.8 \\
FREQ$^\diamond$ \cite{neural-motif} &  &  &  & 4.5 &  & 6.1 &  & 7.1 & 5.1 &  & 7.2 &  & 8.5 & 8.3 &  & 13.0 &  & 16.0& 9.8 \\
MOTIFS$^\diamond$ \cite{neural-motif} &  &  &  & 4.2 &  & 5.7 &  & 6.6 & 6.3 &  & 7.7 &  & 8.2 & 10.8 &  & 14.0 &  & 15.3 & 9.6 \\
KERN$^\diamond$  \cite{kern} &  &  &  & - &  & 6.4 &  & 7.3 &  &  & 9.4 &  & 10.0 & - &  & 17.7 &  & 19.2& 11.7 \\
Chain$^\diamond$ \cite{vctree} &  &  &  & 4.6 &  & 6.3 &  & 7.2 & 6.3 &  & 7.9 &  & 8.8 & 11.0 &  & 14.4 &  & 16.6& 10.2 \\
Overlap$^\diamond$ \cite{vctree} &  &  &  & 4.8 &  & 6.5 &  & 7.5 & 7.2 &  & 9.0 &  & 9.3 & 12.5 &  & 16.1 &  & 17.4 & 11.0\\
Multi-Branch$^\diamond$ \cite{vctree} &  &  &  & 4.7 &  & 6.5 &  & 7.4 & 6.9 &  & 8.6 &  & 9.2 & 11.9 &  & 15.5 &  & 16.9 &10.7 \\
VCTREE-SL$^\diamond$ \cite{vctree} &  &  &  & 5.0 &  & 6.7 &  & 7.7 & 8.0 &  & 9.8 &  & 10.5 & 13.4 &  & 17.0 &  & 18.5 &11.7 \\
VCTREE-HL$^\diamond$ \cite{vctree} &  &  &  & 5.2 &  & 6.9 &  & 8.0 & 8.2 &  & 10.1 &  & 10.8 & 14.0 &  & 17.9 &  & 19.4 &12.2 \\ \hline
\bf GPS-Net$^\diamond$ &  &  &  & \bf6.9 &  &\bf 8.7 &  &\bf 9.8 &\bf 10.0 &  & \bf11.8 &  &\bf 12.6 & \bf17.4 &  &\bf 21.3 &  & \bf22.8&  \bf 14.5\\ \hline
\end{tabular}}
{\topcaption{Mean recall $(\%)$ of various methods across all the 50 relationship categories. All methods in this table adopt the same Faster-RCNN detector.}\label{tba5}}
\end{table*}

\subsection{Performance Comparison on VRD with various $k$}
\begin{table*}[h]
\setlength\abovecaptionskip{-0.2\baselineskip}
\setlength\belowcaptionskip{3pt}
\setlength{\tabcolsep}{1.0mm}{
\begin{tabular}{l|l|ccc|ccccccc|ccccccc}
\hline
 & & \multicolumn{3}{c|}{Pre.} & \multicolumn{7}{c|}{Rel.} & \multicolumn{7}{c}{Phr.} \\
& & \multicolumn{3}{c|}{k=1} & \multicolumn{3}{c}{k=1} &  & \multicolumn{3}{c|}{k=70} & \multicolumn{3}{c}{k=1} &  & \multicolumn{3}{c}{k=70} \\
\multicolumn{1}{c|}{Pretrained} & Model & R@50 &  & R@100 & R@50 &  & R@100 &  & R@50 &  & R@100 & R@50 &  & R@100 &  & R@50 &  & R@100 \\ \hline \hline
&VRD-Full \cite{vrd} & 47.9 &  & 47.9 & 16.2 &  & 17.0 &  & - &  & - & 13.9 &  & 14.7&  & - &  & - \\
&PPRFCN \cite{PPRFCN} & 47.4 &  & 47.4 & 19.6 &  & 23.2 &  & - &  & - & 14.4 &  & 15.7&  & - &  & - \\
&VTranse \cite{Vtranse} & 44.8 &  & 44.8 & 19.4 &  & 22.4 &  & - &  & - & 14.1 &  & 15.2&  & - &  & - \\
&Vip-CNN \cite{ViP-CNN}& - &  & - & 17.3 &  & 20.0 &  & - &  & - & 22.8 &  & 27.9 &  & - &  & - \\
\multicolumn{1}{c|}{Unknown}& VRL \cite{VRL}  & - &  & - & 18.2 &  & 20.8 &  & - &  & - & 21.4 &  & 22.6 &  & - &  & - \\
&KL distilation\cite{KL} & 55.2 &  & 55.2 & 19.2 &  & 21.3 &  & 22.7&  & 31.9 & 23.1 &  & 24.0 &  & 26.3 &  & 29.4 \\
&MF-URLN \cite{zhan}& 58.2 &  & 58.2 & 23.9 &  & 26.8 &  & - &  & - & 31.5 &  & 36.1 &  & - &  & - \\ \hline
&Zoom-Net \cite{ZOOM-NET}& 50.7 &  & 50.7 & 18.9 &  & 21.4 &  &  21.4 & & 27.3 & 24.8 &  & 28.1 &  & 29.1 &  & 37.3 \\
\multicolumn{1}{c|}{\multirow{2}{*}{ImageNet}}&CAI+SCA-M \cite{ZOOM-NET}& 56.0 &  & 56.0 & 19.5 &  & 22.4 &  & 22.3 &  & 28.5 & 25.2 &  & 28.9 &  & 29.6 &  & 18.4 \\
&RelDN\cite{contrastive} & - &  & - & 19.8 &  & 23.0 &  & 21.5 &  & 26.4 & 26.4 &  & 31.4 &  & 28.2 &  & 25.4 \\
&{\bf GPS-Net} & \bf58.7 &  & \bf58.7 &\bf 21.5 &  & \bf24.3 &  & \bf23.6 &  & \bf28.9 &\bf 28.9 &  & \bf34.0 &  & \bf30.4 &  &\bf 38.2 \\ \hline
\multicolumn{1}{c|}{\multirow{2}{*}{COCO}}&RelDN \cite{contrastive} & - &  & - & 25.3 &  & 28.6 &  & 28.2 &  & 33.9 & 31.3 &  & 36.4 &  & 34.5 &  & 42.1 \\
&{\bf GPS-Net} & \bf 63.4 &  & \bf63.4 & \bf27.8 &  &\bf 31.7 &  &\bf 30.6 &  & \bf37.0 & \bf33.8 &  &\bf 39.2 &  & \bf36.8 &  &\bf 44.5 \\ \hline
\end{tabular}}
{\topcaption{Performance comparison with state-of-the-art methods on the VRD dataset. Pre., Phr., and Rel. represent predicate detection, phrase detection, and relation detection, respectively. $-$ denotes that the result is unavailable.}\label{tabvv}}
\end{table*}
As revealed in previous works \cite{ZOOM-NET, KL, contrastive}, each object pair may be described by several plausible predicates. In other words, it should have been formulated as a multi-label classification problem. Therefore, evaluation metrics based on the top-1 prediction ($k$=1) per object pair only may be unreasonable. Following \cite{ZOOM-NET, KL, contrastive}, we further report recall with respect to different $k$ ($k$=1, 70) and compare with state-of-the-art methods. As shown in Table \ref{tabvv}, GPS-Net consistently achieves the best performance among state-of-the-art methods.

\section{Qualitative Analysis}\label{qualiti}
\subsection{Generated Scene Graph}
\begin{figure*}[h]
\begin{center}
\includegraphics[scale=0.082]{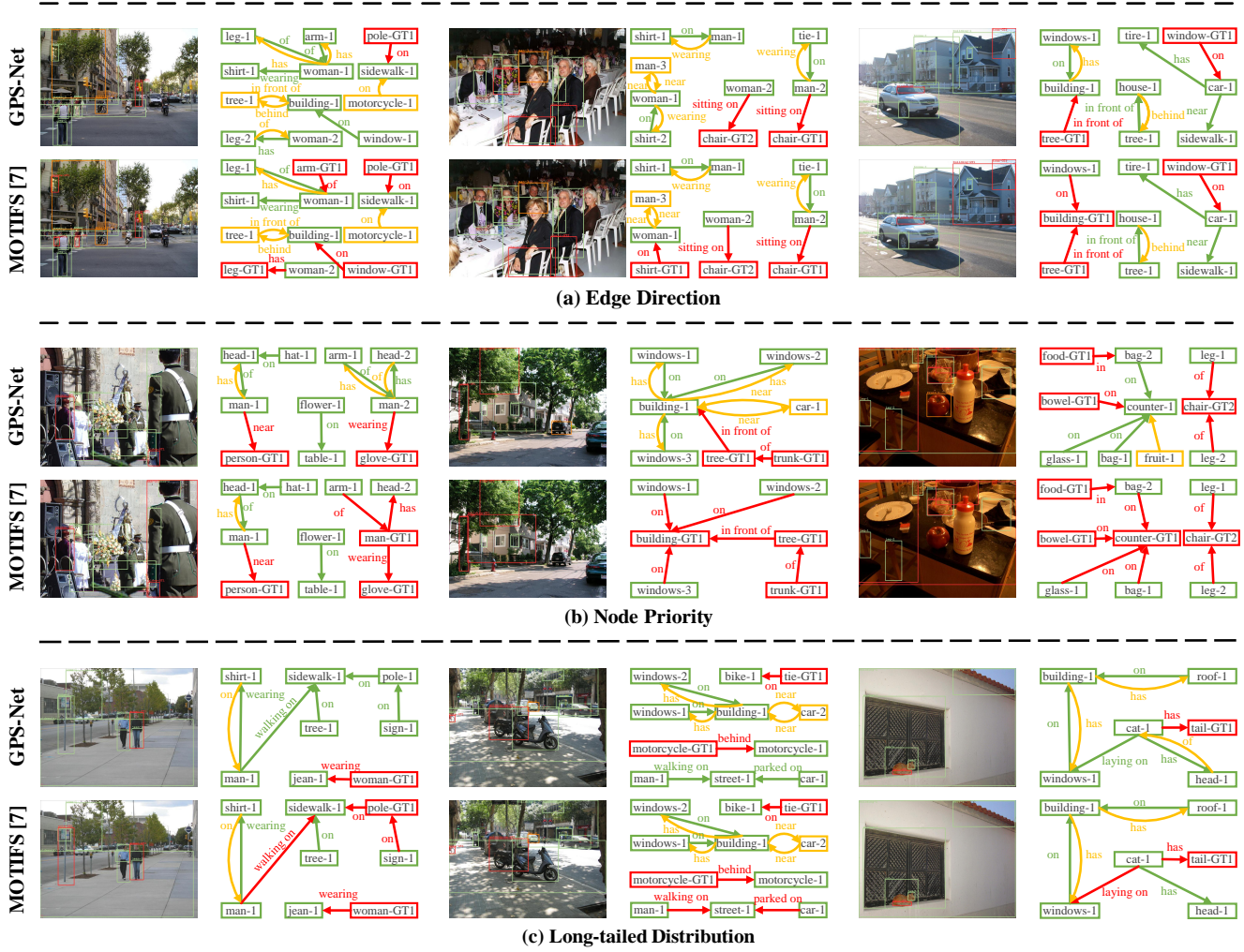}
\caption{Qualitative comparisons between GPS-Net and MOTIF with R@20 in the SGDET setting. Green boxes are detected objects with IOU larger than 0.5 with the ground-truth. Green edges are predictions of relationships that are consistent with the ground-truth. Yellow boxes (edges) denote reasonable detections of objects (relationships), but are not annotated in the database. Red boxes (edges) represent ground-truth objects (relationships) that have no match with the detection results by the algorithm.}
\label{qualitative}
\end{center}
\end{figure*}
Figure \ref{qualitative} illustrates qualitative comparisons between GPS-Net and MOTIFS \cite{neural-motif}. In Figure \ref{qualitative}(a), it is shown that for nodes with low priority and relationships with high frequency, GPS-Net still makes better predictions than MOTIFS. Therefore, we owe this performance gain to the DMP module that encodes edge direction information and provide node-specific context. In Figure \ref{qualitative}(b), it is shown that GPS-Net makes fewer mistakes than MOTIFS for nodes of high priority. We give this credit to the NPS-loss. Finally, in Figure \ref{qualitative}(c), it can be observed that GPS-Net makes outstanding improvement in predicting low-frequency relationships, \textit{e.g}., \textit {walking on} and \textit {wearing}, via the help of the ARM module.

\subsection{Attention Maps of Different MP Modules}

\begin{figure*}[h]
\begin{center}
\includegraphics[scale=0.53]{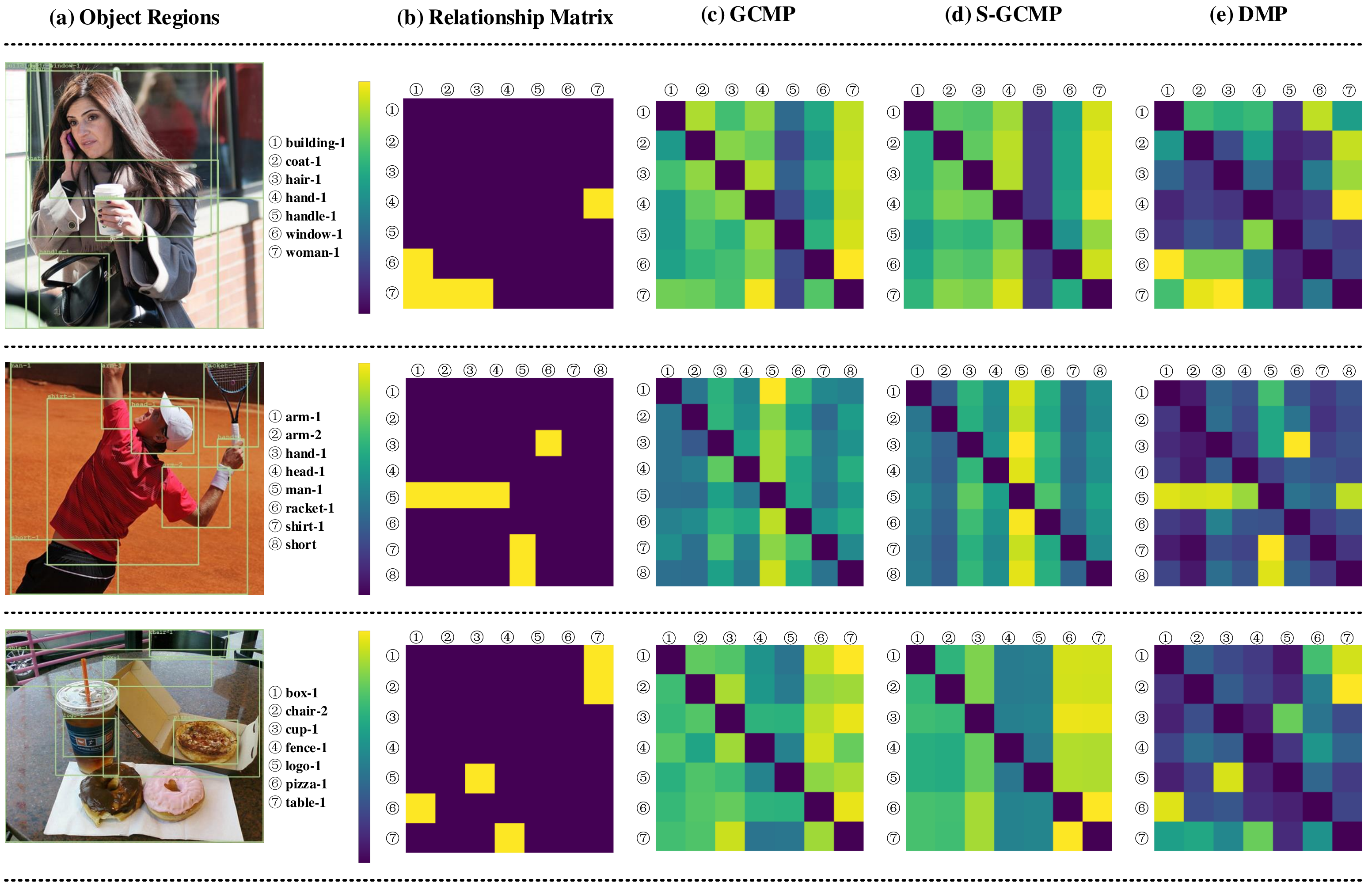}
\par\nointerlineskip\vspace{-1mm}
\caption{Attention Maps of Different MP Modules.}
\label{atten}
\end{center}
\end{figure*}

We make qualitative comparisons between GCMP, S-GCMP, and DMP in Figure \ref{atten}. Figures \ref{atten}(a) and (b) show ground-truth object regions and the ground-truth relationship matrix. More specifically, in the relationship matrix, yellow cube denotes one relationship is presented, and purple cube represents the opposite. Figures \ref{atten}(c)(d)(e) show the attention maps produced by GCMP, S-GCMP, and DMP respectively. It is clear that GCMP and S-GCMP produce very similar context for each node (elements in each column are similar). Only DMP obtains node-specific context (elements in each column are diverse). Furthermore, the attention map produced by DMP is highly consistent with the ground-truth relationship matrix in Figures \ref{atten}(b). Therefore, our proposed DMP module plays a key role in SGG, helping GPS-Net to achieve state-of-the-art performance.

\end{appendices}


\begin{thebibliography}{l}
\bibitem{MSDN}\label{2}
Y. Li, W. Ouyang, B. Zhou, K. Wang, and X. Wang. Scene graph generation from objects, phrases and region captions. In {\it{ICCV}}, 2017.

\bibitem{vctree}\label{3}
K. Tang, H. Zhang, B. Wu, W. Luo, and W. Liu. Learning to compose dynamic tree structures for visual contexts. In {\it{CVPR}}, 2019.


\bibitem{3D}\label{4}
S. Qi, Y. Zhu, S. Huang, C. Jiang, S. Zhu. Human-centric Indoor Scene Synthesis Using Stochastic Grammar. In {\it{ICLR}}, 2018.

\bibitem{scenegraph}\label{1}
J. Johnson, R. Krishna, M. Stark, L. Li, D. Shamma, M. Bernstein, and L. Fei-Fei. Image retrieval using scene graphs. In {\it{CVPR}}, 2015.


\bibitem{imp}\label{5}
D. Xu, Y. Zhu, C. Choy, and L. Fei-Fei. Scene graph generation by iterative message passing. In {\it{CVPR}}, 2017.

\bibitem{contrastive}\label{6}
J. Zhang, K. Shih, A. Elgammal, A. Tao, and B. Catanzaro. Graphical contrastive losses for scene graph parsing. In {\it{CVPR}}, 2019.

\bibitem{neural-motif}\label{7}
R. Zellers, M. Yatskar, S. Thomson, and Y. Choi. Neural motifs: Scene graph parsing with global context. In {\it{CVPR}}, 2018.

\bibitem{gat}\label{8}
P. Velickovic, G. Cucurull, A. Casanova, and A. Romero. Graph Attention Networks. In {\it{ICLR}}, 2018.

\bibitem{direction}\label{9}
L. Gong and Q. Cheng. Exploiting Edge Features in Graph Neural Networks. In {\it{CVPR}}, 2019.

\bibitem{attentive}\label{10}
M. Qi, W. Li, Z. Yang, Y. Wang and J. Luo. Attentive Relational Networks for Mapping Images to Scene Graphs. In {\it{CVPR}}, 2019.

\bibitem{agent}\label{11}
L. Chen, H. Zhang, J. Xiao, X. He, S. Pu, and S. F. Chang, Counterfactual Critic Multi-Agent Training for Scene Graph Generation. In {\it{CVPR}}, 2019.

\bibitem{mutan}\label{12}
H. Ben-Younes, R. Cadene, M. Cord, and N. Thome. Mutan: Multimodal tucker fusion for visual question answering. In {\it{CVPR}}, 2017.

\bibitem{focalloss}\label{13}
T.-Y. Lin, P. Goyal, R. Girshick, K. He, and P. Dollar. Focal loss for dense object detection. In {\it{ICCV}}, 2017.

\bibitem{VGdataset}\label{14}
R. Krishna, Y. Zhu, O. Groth, J. Johnson, K. Hata, J. Kravitz, S. Chen, Y. Kalantidis, L.-J. Li, D. A. Shamma, et al. Visual genome: Connecting language and vision using crowdsourced dense image annotations. In {\it{IJCV}}, 2017.

\bibitem{OI}\label{15}
A. Kuznetsova, H. Rom, N. Alldrin, J. Uijlings, I. Krasin, J. Tuset, S. Kamali, S. Popov, M. Malloci, T. Duerig, et al. The open imagesdataset v4: Unified image classification, object detection, and visual relationship detection at scale. In {\it{arXiv:1811.00982}}, 2018.

\bibitem{vrd}\label{16}
C. Lu, R. Krishna, M. Bernstein, and L. Fei-Fei. Visual relationship detection with language priors. In {\it{ECCV}}, 2016.

\bibitem{senet}
J. Hu, L. Shen, and G. Sun. Squeeze-and-excitation networks. In {\it{CVPR}}, 2018.

\bibitem{psanet}
H. Zhao, Y. Zhang, S. Liu, J. Shi, C. Change Loy, D. Lin, and J. Jia. Psanet: Point-wise spatial attention network for scene parsing. In {\it{ECCV}}, 2018.

\bibitem{nlnet}
X.Wang, R. Girshick, A. Gupta, and K. He. Non-local neural networks. In {\it{CVPR}}, 2018.

\bibitem{ccnet}
Z. Huang, X. Wang, L. Huang, C. Huang, Y. Wei, and W. Liu. Ccnet: Criss-cross attention for semantic seg- mentation. In {\it{ arXiv preprint arXiv:1811.11721}}, 2018.

\bibitem{gcnet}\label{21}
Y. Cao and J. Xu. GCNet: Non-local Networks Meet Squeeze-Excitation Networks and Beyond. In {\it{arXiv preprint arXiv:1904.11492}}, 2019.

\bibitem{graphrcnn}\label{22}
J. Yang, J. Lu, S. Lee, D. Batra, and D. Parikh. Graph r-cnn for scene graph generation. In {\it{ECCV}}, 2018.

\bibitem{kern}\label{23}
T. Chen, W. Yu, R. Chen, and L. Lin. Knowledge-embedded routing network for scene graph generation. In {\it{CVPR}}, 2019.

\bibitem{gpi}\label{16}
R. Herzig, M. Raboh, G. Chechik, J. Berant, and A. Globerson. Mapping images to scene graphs with permutation-invariant structured prediction. In {\it{NeurIPS}}, 2018.

\bibitem{fasterrcnn}\label{24}
S. Ren, K. He, R. Girshick, and J. Sun. Faster r-cnn: Towards real-time object detection with region proposal networks. In {\it{NIPS}}, 2015.

\bibitem{px2graph}\label{25}
A. Newell and J. Deng. Pixels to graphs by associative embedding. In {\it{NIPS}}, 2017.

\bibitem{tfr}\label{14}
S. Hwang, S. Ravi, Z. Tao, H. Kim, M. Collins, and V. Singh. Tensorize, factorize and regularize: Robust visual relationship learning. In {\it{CVPR}}, 2018.


\bibitem{largescale}\label{17}
J. Zhang, Y. Kalantidis, M. Rohrbach, M. Paluri, A. Elgammal, and M. Elhoseiny. Large-scale visual relationship un- derstanding. In {\it{AAAI}}, 2019.

\bibitem{maskrcnn}\label{19}
K. He, G. Gkioxari, P. Doll\'{a}r, and R. Girshick. Mask r-cnn. In {\it{ICCV}}, 2017.

\bibitem{factorize}\label{20}
Y. Li, W. Ouyang, B. Zhou, Y. Cui, J. Shi, and X. Wang. Factorizable net: An efficient subgraph-based framework for scene graph generation. In {\it{ECCV}}, 2018.

\bibitem{sgandcaption}\label{21}
Y. Li, W. Ouyang, B. Zhou, K. Wang, and X. Wang. Scene graph generation from objects, phrases and region captions. In {\it{ICCV}}, 2017.

\bibitem{exploringedge}\label{23}
L. Gong, and Q. Cheng. Exploiting Edge Features for Graph Neural Networks. In {\it{CVPR}}, 2019.

\bibitem{LN}\label{24}
J. Ba, J. R. Kiros, and G. E. Hinton, Layer normalization. In {\it{arXiv preprint arXiv:1607.06450}}, 2016.

\bibitem{MLB}\label{26}
J. Kim, K. On, W. Lim, J. Kim, J. Ha, and B. Zhang. Hadamard product for low-rank bilinear pooling. In {\it{ICLR}}, 2017.

\bibitem{PPRFCN}\label{27}
H. Zhang, Z. Kyaw, J. Yu, and S.-F. Chang. Ppr-fcn: Weakly supervised visual relation detection via parallel pairwise r-fcn. In {\it{CVPR}}, 2017.

\bibitem{SA-FULL}
J. Peyre, I. Laptev, C. Schmid, and J. Sivic. Weakly-supervised learning of visual relations. In {\it{ICCV}}, 2017.

\bibitem{Vtranse}
H. Zhang, Z. Kyaw, S. Chang, and T. Chua. Visual translation embedding network for visual relation detection. In {\it{CVPR}}, 2017.

\bibitem{DR-NET}
B. Dai, Y. Zhang, and D. Lin. Detecting visual relationships with deep relational networks. In {\it{CVPR}}, 2017.

\bibitem{ViP-CNN}
Y. Li, W. Ouyang, and X. Wang. Vip-cnn: A visual phrase reasoning convolutional neural network for visual relationship detection. In {\it{CVPR}}, 2017.

\bibitem{VRL}
X. Liang, L. Lee, and E. P. Xing. Deep variation-structured reinforcement learning for visual relationship and attribute detection. In {\it{CVPR}}, 2017.

\bibitem{CAI}
B. Zhuang, L. Liu, C. Shen, and I. Reid. Towards context-aware interaction recognition for visual relationship detec- tion. In {\it{ICCV}}, 2017.

\bibitem{ZOOM-NET}
G. Yin, L. Sheng, B. Liu, N. Yu, X. Wang, J. Shao, and C. Change Loy. Zoom-net: Mining deep feature interactions for visual relationship recognition. In {\it{ECCV}}, 2018.

\bibitem{KL}
R. Yu, A. Li, V. I. Morariu, and L. S. Davis. Visual relationship detection with internal and external linguistic knowl- edge distillation. In {\it{ICCV}}, 2017.

\bibitem{zhan}
Y. Zhan, J. Yu, T. Yu, D. Tao. On Exploring Undetermined Relationships for Visual Relationship Detection. In {\it{CVPR}}, 2019.

\bibitem{GCN}
T. N. Kipf and M. Welling. Semi-Supervised Classification with Graph Convolutional Networks. In {\it{ICLR}}, 2017.


\bibitem{evolved}
 R. Houthooft, R. Y. Chen, P. Isola, B. C. Stadie, F. Wolski, J. Ho, and P. Abbeel. Evolved policy gradients.
 In {\it{NeurlPS}}, 2018.



\bibitem{fusion}
Y. Zhang, J. Hare, and A. Prugel-Bennett. Learning to count objects in natural images for visual question answering. In {\it{ICLR}}, 2018.



\end{thebibliography}
\end{document}